\documentclass[10pt]{article}

\usepackage[utf8]{inputenc}
\usepackage[T1]{fontenc}
\usepackage{epsfig}
\usepackage{graphicx}
\usepackage[inkscapelatex=false]{svg}
\usepackage{algorithm}
\usepackage{algorithmic}
\usepackage{mathrsfs}
\usepackage{amsfonts}
\usepackage{float}
\usepackage{amssymb,amsmath}
\usepackage{indentfirst}
\usepackage{bm}
\usepackage{dsfont}
\usepackage{makecell}
\usepackage{theorem}
\usepackage{multirow}
\usepackage{array}
\usepackage{multirow}
\usepackage{subfigure}
\usepackage{longtable}
\usepackage{threeparttable}
\usepackage{booktabs}
\usepackage{todonotes}

\newif\iftwoc
\twocfalse

\newif\ifpfig
\pfigtrue

\iftwoc

\makeatletter
\def\@normalsize{\@setsize\normalsize{12pt}\xpt\@xpt
\abovedisplayskip 12pt plus3pt minus7pt\belowdisplayskip \abovedisplayskip
\abovedisplayshortskip \z@ plus3pt\belowdisplayshortskip 6.5pt plus3.5pt
minus3pt\let\@listi\@listI}

\def\subsize{\@setsize\subsize{12pt}\xipt\@xipt}

\def\section{\@startsection {section}{1}{\z@}{24pt plus 2pt minus 2pt}
{12pt plus 2pt minus 2pt}{\large\bf}}

\def\subsection{\@startsection {subsection}{2}{\z@}{12pt plus 2pt minus 2pt}
{12pt plus 2pt minus 2pt}{\subsize\bf}}
\makeatother

\else

\renewcommand{\baselinestretch}{1.2}

\fi

\newlength{\wone}
\newcommand{\eq}{\begin{equation}}
\newcommand{\en}{\end{equation}}

\begin{document}
\date{}
\title{\Large\bf PolySmart @ TRECVid 2024 Medical Video Question Answering}
\author{
	Jiaxin Wu$^\dagger$, Yiyang Jiang$^\dagger$, Xiao-Yong Wei$^{\star\dagger}$, Qing Li$^\dagger$ 
    \vspace{0.08in}
   \\ {\em $^\dagger$Department of Computing, The Hong Kong Polytechnic University}
     \\    {\em $^\star$Department of Computer Science, Sichuan University}
   \\ jiaxwu@polyu.edu.hk, yiyang.jiang@polyu.edu.hk, \\ cs007.wei@polyu.edu.hk, prof.li@polyu.edu.hk \\
}
\maketitle

\section*{\centering Abstract}
\indent In this paper, we summarize our submitted runs and results for the Medical Video Question Answering task at TRECVid 2024\cite{2023trecvidawad}.

\vspace{0.2cm}
\label{abs:avs}\noindent\textbf{Video Corpus Visual Answer Localization (VCVAL):} 
This task includes question-related video retrieval and visual answer localization in the videos. Specifically, we use text-to-text retrieval to find relevant videos for a medical question based on the similarity of video transcript and answers generated by GPT4. For the visual answer localization, the start and end timestamps of the answer are predicted by the alignments on both visual content and subtitles with queries. We submit five runs this year and they are briefly summarized  as follows:
\begin{itemize}
\item[\textbullet]\textit{Run 1}: Achieves MAP $= 0.1401$ using top-10 text-to-text retrieval. This run computes the similarity between the original question and the video transcript using two sentence-transformer models: \texttt{PubMedBert} and \texttt{MiniLM}.

\item[\textbullet]\textit{Run 2}: Achieves MAP $= 0.1305$ with top-10 text-to-text retrieval. This run evaluates the similarity between GPT-4 generated question-answer pairs and the video transcript, using the same sentence-transformer models as in Run 1.

\item[\textbullet]\textit{Run 3}: Achieves MAP $= 0.1348$ for top-100 text-to-text retrieval. This run combines the results of Run 1 and Run 2.

\item[\textbullet]\textit{Run 4}: Achieves MAP $= 0.1087$ with top-10 text-to-text retrieval. This run takes the mean similarity between the original question and the video transcript, using the same sentence-transformer models as in Run 1.

\item[\textbullet]\textit{Run 5}: A novel approach using top-100 text-to-vision retrieval with BLIP-2 features, achieving MAP $= 0.0466$.
\end{itemize}

\label{abs:avs}\noindent\textbf{Query-Focused Instructional Step Captioning (QFISC)}: 
For this task, the step captions are generated by GPT4. Specifically, we provide the video captions generated by the LLaVA-Next-Video model and the video subtitles with timestamps as context, and ask GPT4 to generate step captions for the given medical query. We only submit one run for evaluation and it obtains a F-score of 11.92 and mean IoU of 9.6527.


\section{Video Corpus Visual Answer Localization}
Medical Video Question Answer Localization (VCVAL) presents unique challenges compared to general text-to-video retrieval \cite{vireo2023,vireo2022,vireo2021,chongwahngo2010trecvid,chongwahngo2008trecvid,chongwahngo2005trecvid} due to the specialized nature of medical content. Unlike general videos, medical videos convey critical information through both precise visual details (e.g., procedures, anatomy) and specific medical terminology, including abbreviations that may not match directly with the video content. This can lead to difficulties in retrieval if the model fails to recognize medical abbreviations in the query, reducing retrieval accuracy. To address these challenges, we introduce a two-step approach combining video retrieval and precise segment localization.

Given the multimodal nature of medical videos, accurately locating answers requires both identifying relevant videos and pinpointing specific segments. Our method performs text-to-text retrieval using video transcripts, obtained via the YouTube API, with sentence transformer embeddings for the query and transcript text. Cosine similarity ranks the results, retrieving top-10 or top-100 videos. In the second stage, a dual-predictor~\cite{weng2022visualanswerlocalizationcrossmodal} system—comprising a Textual Predictor and a Visual Predictor—focuses on complementary aspects of video content to refine segment localization. A cross-modal knowledge transfer mechanism with a lookup table facilitates information exchange between predictors, enabling adaptive knowledge sharing. This system captures both visual and textual nuances, enhancing answer localization accuracy. An Optimized Dynamic Learning (ODL) module adjusts knowledge transfer based on each predictor’s needs, further improving robustness across varying scenarios.

\subsection{Related Video Retrieval by Text}
In the first step of our approach, we perform text-to-text retrieval to identify videos relevant to the medical query. This step reduces the search space by retrieving a subset of the most relevant videos before moving to the finer task of segment localization.
\subsubsection{Video Transcript Extraction and Question Expansion}
For each video \( V_i \) in the corpus, we use the YouTube API to extract its transcript, denoted as \( T_i \). This transcript represents the spoken content within the video.

To enhance the query’s richness and handle medical terminology, we generate an additional embedding based on a GPT-4-enhanced query answer \( q_{\text{GPT-4}} \). The prompt for GPT-4 is:

\begin{quote}
\textit{"You act as a medical or a health helper. Given a list of medical or health-related how-to questions, output the instructions step by step."}
\end{quote}

\subsubsection{Text Feature Extraction and Alignment}

We utilize a sentence transformer model~\cite{li2023blip2bootstrappinglanguageimagepretraining,reimers-2019-sentence-bert} to encode both the query and the transcripts into vector representations in a semantic embedding space. Given a query \( q \) and a video transcript \( T_i \), we compute the embeddings as follows:
   
   \[
   \mathbf{q}_{\text{orig}} = \text{SentenceTransformer}(q)
   \]
   \[
   \mathbf{t}_i = \text{SentenceTransformer}(T_i)
   \]
   \[
   \mathbf{q}_{\text{GPT-4}} = \text{SentenceTransformer}(q_{\text{GPT-4}})
   \]

We compute the cosine similarity between the embeddings of each video transcript \( \mathbf{t}_i \) and both the original query embedding \( \mathbf{q}_{\text{orig}} \) and the GPT-4-enhanced query embedding \( \mathbf{q}_{\text{GPT-4}} \). The cosine similarity for each query embedding with respect to \( T_i \) is defined as:

   \[
   \text{Sim}(\mathbf{q}_{\text{orig}}, \mathbf{t}_i) = \frac{\mathbf{q}_{\text{orig}} \cdot \mathbf{t}_i}{\|\mathbf{q}_{\text{orig}}\| \|\mathbf{t}_i\|}
   \]
   \[
   \text{Sim}(\mathbf{q}_{\text{GPT-4}}, \mathbf{t}_i) = \frac{\mathbf{q}_{\text{GPT-4}} \cdot \mathbf{t}_i}{\|\mathbf{q}_{\text{GPT-4}}\| \|\mathbf{t}_i\|}
   \]

   The final similarity score for each video \( V_i \) is the maximum of the two similarity values:

   \[
   \text{Sim}_{\text{final}}(q, T_i) = \max \left( \text{Sim}(\mathbf{q}_{\text{orig}}, \mathbf{t}_i), \text{Sim}(\mathbf{q}_{\text{GPT-4}}, \mathbf{t}_i) \right)
   \]

Based on the computed similarity scores \( \text{Sim}_{\text{final}}(q, T_i) \) for each video transcript \( T_i \), we rank the videos and retrieve the top-10 or top-100 videos with the highest scores. These selected videos are then passed to the subsequent localization stage.

\subsection{Visual Answer Localization by Multi-Modal Collaboration}

\begin{figure}[t!]
\includegraphics[width=1\linewidth]{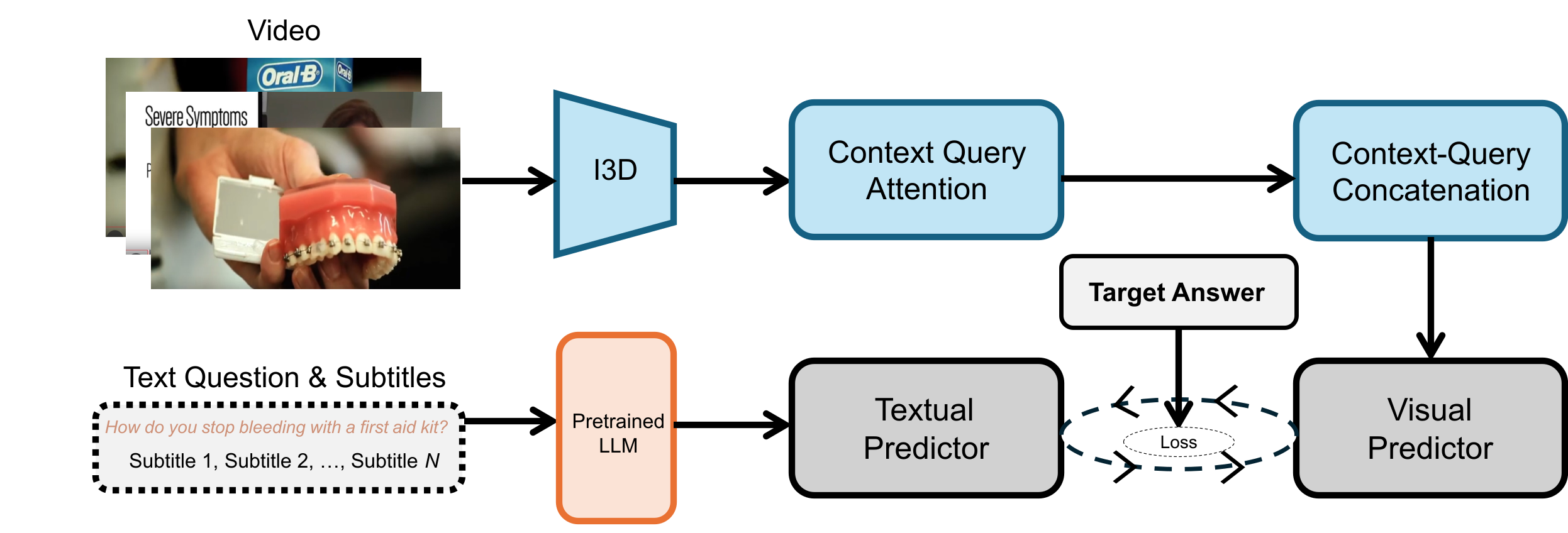}
    \caption{Multi-Modal Collaborative Localization}
    \label{fig:1}
\end{figure}

In tackling the Video Corpus Visual Answer Localization (VCVAL) task, we introduce Multi-Modal Collaborative Localization (Figure~\ref{fig:1}), which employs a synergistic approach to cross-modal learning by integrating feature extraction, cross-modal fusion, and adaptive knowledge transfer.

\subsubsection{Feature Extraction}

We utilize the I3D~\cite{8099985} model, a powerful network for video processing, to encode the visual information from each video segment. I3D operates by inflating 2D convolutional filters into 3D, allowing it to effectively capture both spatial and temporal features. This model generates a feature matrix \( V \in \mathbb{R}^{k \times d} \), where \( k \) is the number of video frames and \( d \) represents the feature dimension. By using I3D, we gain high-quality visual representations that capture both static spatial information and dynamic motion patterns crucial for video answer localization.

For the textual modality, we employ the DeBERT~\cite{he2023debertav3improvingdebertausing} to process the concatenated text, including the query \( Q \) and any relevant video captions \( T = [Q, T_1, ..., T_r] \). This produces a feature vector \( T \in \mathbb{R}^{n \times d} \), where \( n \) is the length of the concatenated text tokens, representing the textual information in alignment with the video content.

\subsubsection{Cross-Modal Fusion}

The MCL approach applies Context Query Attention (CQA)~\cite{zhang2020spanbasedlocalizingnetworknatural} to merge visual and textual features effectively. CQA leverages two attention mechanisms: query-to-context and context-to-query, facilitating deeper semantic interaction between the modalities. This fusion step results in enhanced feature representations, enabling better alignment between the video content and the query.

\subsubsection{Dual Predictors for Localization}

To determine the start and end of the visual answer, MCL utilizes two predictors: a Visual Predictor and a Textual Predictor. The Visual Predictor employs LSTMs followed by feedforward networks to identify key time points in the video, while the Textual Predictor, based on the structure of question-answering networks, predicts relevant time spans from textual features. This dual predictor setup ensures robust answer localization by leveraging insights from both modalities.

\subsubsection{Adaptive Knowledge Transfer}

A critical component of MCL is the Adaptive Knowledge Transfer Module. To synchronize the predictions of the Visual and Textual Predictors, we introduce a Lookup Table that facilitates cross-modal knowledge alignment. By mapping predicted time spans from one modality to another, the Lookup Table ensures consistent understanding across the predictors.

MCL employs a One-Way Dynamic Loss Adjustment mechanism from previous work~\cite{weng2022visualanswerlocalizationcrossmodal}, dynamically adjusting the knowledge transfer between modalities based on prediction alignment using an Intersection over Union (IoU) criterion. This process stops gradient flow between predictors, enabling them to independently refine their learning while optimizing for overall consistency. The final loss function combines contributions from each predictor and includes additional loss terms from cross-modal transfer. The total loss is defined as:

\[
\text{Loss} = \text{Loss}_{\text{Visual}} + \text{Loss}_{\text{Textual}} + \text{Loss}_{\text{Transfer}}^{\text{Visual}} + \text{Loss}_{\text{Transfer}}^{\text{Textual}}
\]

\section{Query-Focused Instructional Step Captioning}
The Query-Focused Instructional Step Captioning (QFISC) task aims to provide step-by-step textual summaries of visual instructional segments within medical videos in response to specific queries. This task extends the visual answer localization approach by requiring the identification of instructional boundaries and the generation of detailed captions for each instructional step, resulting in a comprehensive response tailored to the medical query.

Using LLaVA NEXT 32B~\cite{zhang2024llavanext-video} with GPT-4~\cite{openai2024gpt4technicalreport}
Our approach begins by using LLaVA NEXT 32B to generate initial captions for each relevant instructional segment in the video. These generated captions are then combined with the original captions from the video, creating a rich dataset that encompasses both generated and existing linguistic cues. This combined data is fed into GPT-4, which processes the information to produce the final output. GPT-4 generates the time ranges for each instructional segment and formulates detailed step-by-step instructions, ensuring that the response aligns accurately with both the visual content and the query requirements.

\subsection{Instructional Video Question Answering}
\begin{figure}[t!]
\includegraphics[width=1\linewidth]{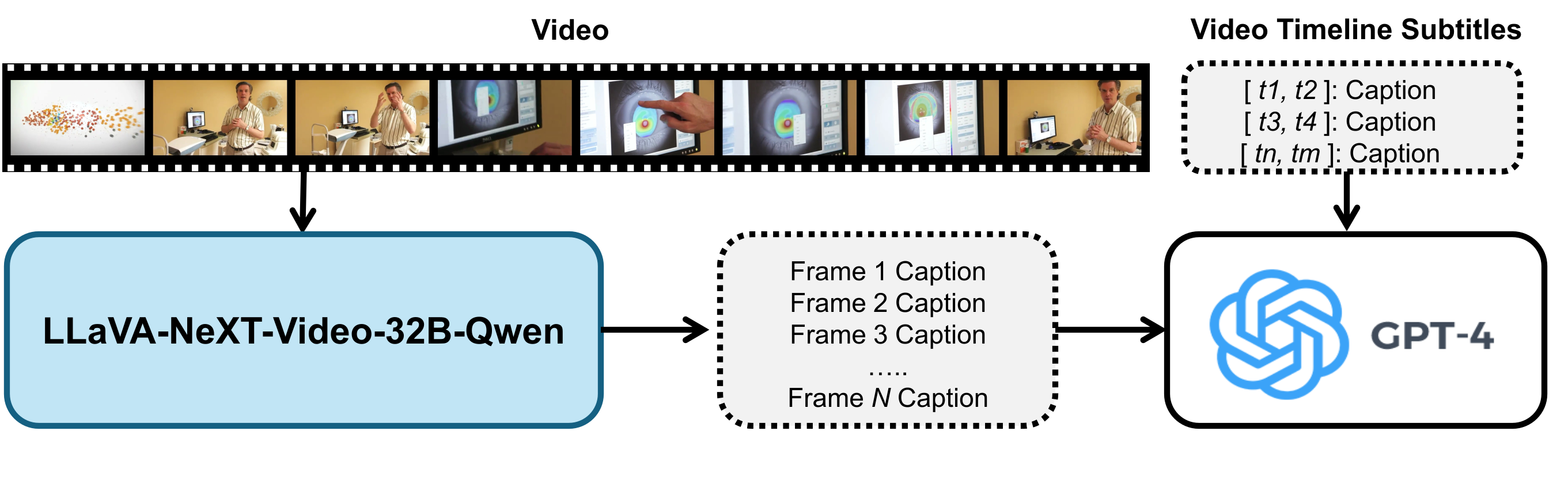}
    \caption{InstructVQA}
    \label{fig:2}
\end{figure}

The InstructVQA model~(Figure~\ref{fig:2}) is designed to tackle the Query-Focused Instructional Step Captioning (QFISC) task by generating structured, step-by-step captions from visual instructional segments in response to a medical query. InstructVQA combines advanced vision-language models with language generation techniques to create temporally-aligned, query-specific instructional summaries.

InstructVQA begins by utilizing LLaVA NEXT 32B~\cite{li2024llavanextinterleavetacklingmultiimagevideo}, a powerful vision-language model, to generate preliminary captions for the relevant instructional segments of a video. Given a medical query \( Q \) and input video \( V \), LLaVA NEXT 32B outputs a set of captions \( C = [c_1, c_2, \dots, c_m] \) that correspond to various instructional steps identified in the video:

\[
C = \text{LLaVA-NEXT-32B}(V, Q)
\]

where \( C \) represents the captions generated based on the visual features of \( V \) aligned with the query \( Q \).

To enhance the depth and accuracy of the generated captions, InstructVQA combines the generated captions \( C \) with the original video subtitles or captions \( S = [s_1, s_2, \dots, s_n] \). This merged caption set, \( C_{\text{combined}} = C \cup S \), incorporates both the generated instructional content and the existing linguistic cues in the video, resulting in a comprehensive set of textual data for the next stage.

The combined captions \( C_{\text{combined}} \) are fed into GPT-4, which processes this text to identify distinct instructional steps. GPT-4 analyzes \( C_{\text{combined}} \) to determine the time range and description for each step, providing a temporally structured and semantically rich response to the query \( Q \). For each step \( i \), GPT-4 outputs a tuple containing the start and end times \( [t_{s,i}, t_{e,i}] \) and a descriptive caption \( d_i \), formulated as follows:

\[
\{(t_{s,i}, t_{e,i}, d_i)\}_{i=1}^p = \text{GPT-4}(C_{\text{combined}})
\]

where \( p \) is the number of instructional steps detected. This allows InstructVQA to produce a sequence of time-aligned, step-by-step instructions.

The final output from InstructVQA is a structured sequence of instructional steps, each aligned with a specific time range and detailed caption, forming a coherent and query-focused instructional guide. Each step \( (t_{s,i}, t_{e,i}, d_i) \) directly corresponds to the query \( Q \), enhancing the usability and relevance of the response.

\section{Results analysis}
\subsection{Video Corpus Visual Answer Localization (VCVAL)}
To evaluate the effectiveness of our proposed method, we conducted experiments on the VCVAL task (Stage of Video retrieval). The results are summarized in Table 1. The performance metrics include Mean Average Precision (MAP), Recall at top 5 (R@5) and top 10 (R@10), Precision at top 5 (P@5) and top 10 (P@10), and normalized Discounted Cumulative Gain (nDCG).

The results Table~\ref{tab:1} demonstrate that our method achieves competitive performance across all metrics. Among the runs, RunID 1 yielded the highest MAP of 0.1401, while RunID 3 also achieved strong results with an MAP of 0.1348. The mean, minimum, and maximum values across the evaluated runs are included for comparison with the reported performance on the VCVAL task.

\begin{table}[h] 
\centering 
\begin{tabular}{cccccccc}
\bottomrule
\textbf{RunID} & \textbf{Model} & \textbf{MAP} & \textbf{R@5} & \textbf{R@10} & \textbf{P@5} & \textbf{P@10} & \textbf{nDCG} \\ 
\bottomrule
1 & $\text{Sim}(\mathbf{q}_{\text{orig}}, \mathbf{t}_i)$&\textbf{0.1401} & \textbf{0.1799} & 0.2094 & 0.1115 & \textbf{0.0635} & 0.1955 \\
2 & $\text{Sim}(\mathbf{q}_{\text{GPT-4}},\mathbf{t}_i)$ & 0.1305 & 0.1767 & \textbf{0.2094} & \textbf{0.1154} & \textbf{0.0635} & 0.1892 \\ 
3 & Run1+ Run2& 0.1348 & 0.1643 & 0.1998 & 0.1077 & 0.0615 & \textbf{0.2009} \\
4 & mean $\text{Sim}(\mathbf{q}_{\text{orig}}, \mathbf{t}_i)$& 0.1087 & 0.1539 & 0.1810 & 0.0885 & 0.0558 & 0.1569 \\
5 & BLIP-2~\cite{li2023blip2bootstrappinglanguageimagepretraining} & 0.0466 & 0.0365 & 0.0756 & 0.0231 & 0.0212 & 0.1167 \\  \hline
\textbf{Min} & & 0.0027 & 0.0027 & 0.0027 & 0.0038 & 0.0019 & 0.0031 \\ 
\textbf{Mean} & & 0.1756 & 0.1972 & 0.2221 & 0.1154 & 0.0649 & 0.2306 \\ 
\textbf{Max} & & 0.4339 & 0.4565 & 0.4857 & 0.2385 & 0.1308 & 0.5443 \\
\bottomrule
\end{tabular}
\label{tab:1}
\caption{Video retrieval results for VCVAL Task} 
\end{table}

\subsubsection{Qualitative result}
\begin{figure}[t!]
\includegraphics[width=1\linewidth]{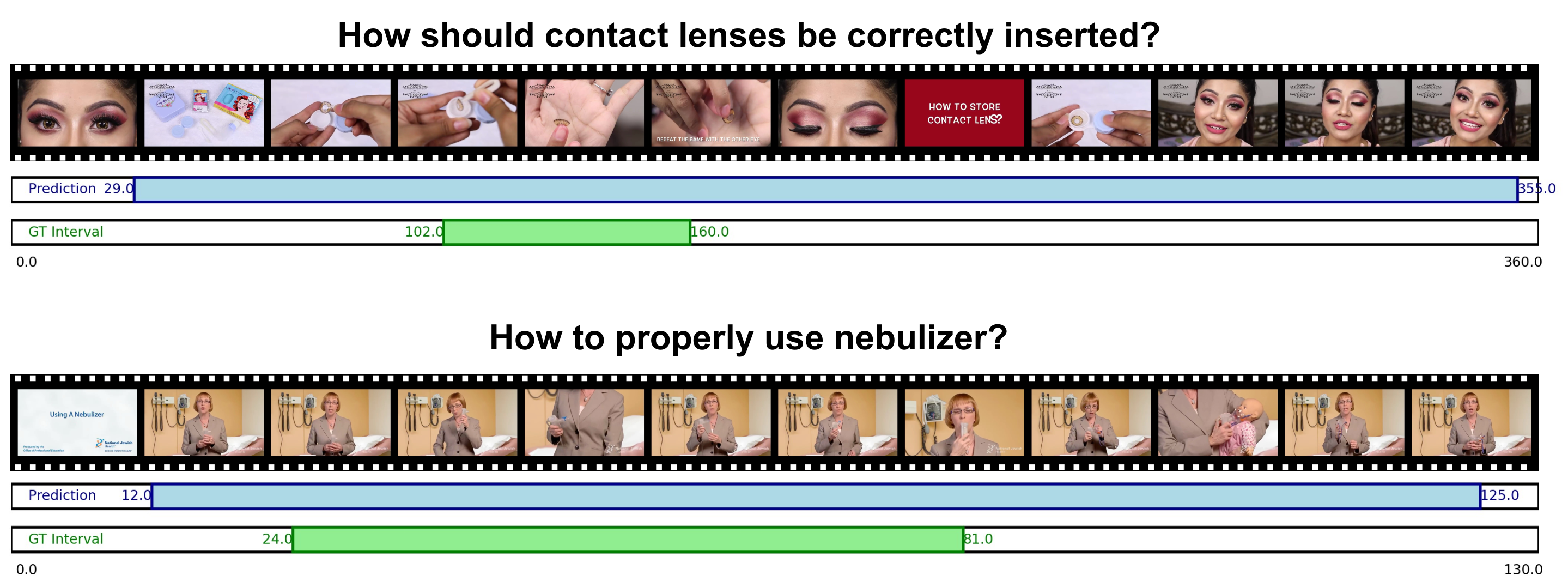}
    \caption{Qualitative result for moment retrieval of VCVAL Task}
    \label{fig:3}
\end{figure}

In our experiments, the model demonstrates strong coverage in retrieving relevant segments within medical videos, though it sometimes lacks precision in identifying exact answer boundaries (Figure~\ref{fig:3}). 

\subsection{Query-Focused Instructional Step Captioning (QFISC)}

\begin{figure}[t!]
\centering
\includegraphics[width=1\linewidth]{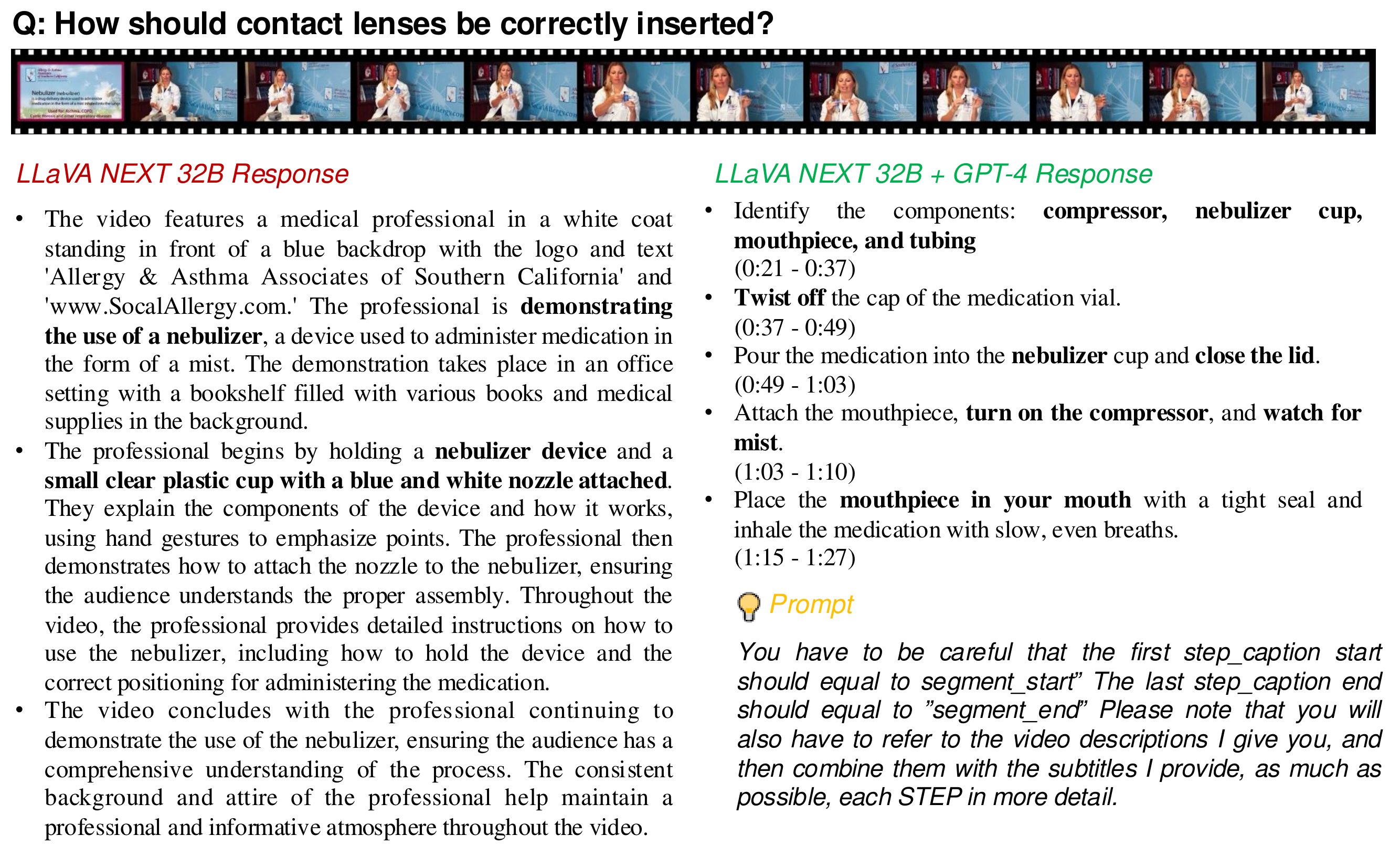}
    \caption{Qualitative result for QFISC}
    \label{fig:4}
\end{figure}

To evaluate our approach for the Query-Focused Instructional Step Captioning (QFISC) task, we measured several metrics, including precision, recall, f-score, overlap IoU (Intersection over Union) at thresholds 3, 5, and 7, as well as the mean IoU (mIoU). The results for the submitted run, as well as the minimum, mean, and maximum values across all runs, are presented in Table~\ref{tab:2}.
\begin{table}[h]
\centering
\begin{tabular}{cccccccc}
\bottomrule
\textbf{Run ID} & \textbf{Precision} & \textbf{Recall} & \textbf{F-Score} & \textbf{IoU@3} & \textbf{IoU@5} & \textbf{IoU@7} & \textbf{mIoU} \\
\bottomrule
1 & 12.5489 & 12.1781 & 11.9291 & 12.1779 & 11.7582 & 8.0083 & 9.6527 \\
\bottomrule
\textbf{Min} & 12.5489 & 10.5014 & 11.9291 & 9.7003 & 9.4781 & 7.3453 & 8.0271 \\ 
\textbf{Mean} & 21.7501 & 25.2405 & 22.1168 & 24.2981 & 22.0943 & 14.1882 & 18.1065 \\
\textbf{Max} & 25.8113 & 35.9927 & 28.7081 & 34.7259 & 32.0150 & 20.0946 & 26.0907 \\
\bottomrule
\end{tabular}
\caption{Experimental Results for Query-Focused Instructional Step Captioning (QFISC) Task}
\label{tab:2}
\end{table}

Our method \texttt{InstructVQA}, demonstrates competitive performance with a precision of 12.5489, recall of 12.1781, and f-score of 11.9291. The overlap IoU metrics indicate robust alignment between the generated captions and the video content, with values across different thresholds.

\subsubsection{Examples of QFISC with LLaVA NEXT 32B and GPT-4}
In our experiments on Query-Focused Instructional Step Captioning (QFISC), we found that using LLaVA NEXT 32B alone did not yield sufficiently structured instructional captions. However, by incorporating GPT-4 to refine and segment the response, we achieved significantly more coherent and detailed step-by-step captions.

\section{Conclusion}
In this notebook, we presented our approach to the TRECVid 2024 Medical Video Question Answering tasks, specifically focusing on Video Corpus Visual Answer Localization (VCVAL) and Query-Focused Instructional Step Captioning (QFISC). By combining a dual-predictor system with cross-modal knowledge transfer and adaptive learning, our method effectively addresses the complexities of medical video localization. For QFISC, the integration of LLaVA NEXT 32B with GPT-4 yielded well-structured, detailed instructional captions that align closely with query requirements. Experimental results demonstrate that our methods provide strong coverage and accuracy, showcasing the potential of multimodal approaches in medical video question answering.

Despite these strengths, our approach faces some limitations. The dual-predictor system sometimes lacks precision in identifying exact segment boundaries, leading to overly broad answer spans. Additionally, the reliance on text-to-text retrieval may struggle with medical abbreviations or terminology that vary across video content, potentially affecting retrieval accuracy. Future work could explore refined localization techniques and enhanced handling of domain-specific language to improve precision and robustness further.

\section{Acknowledgments}
This research project is supported by the National Natural Science Foundation of China (Grant No.: 62372314). 


\bibliographystyle{IEEEtran}
\bibliography{trec}
\end{document}